\documentclass[letterpaper]{article} 
\usepackage{aaai24}  
\usepackage{times}  
\usepackage{helvet}  
\usepackage{courier}  
\usepackage[hyphens]{url}  
\usepackage{graphicx} 
\usepackage{natbib}  
\usepackage{caption} 
\usepackage{algorithm}
\usepackage{algorithmic}

\usepackage{newfloat}
\usepackage{listings}
\DeclareCaptionStyle{ruled}{labelfont=normalfont,labelsep=colon,strut=off} 
\floatstyle{ruled}
\newfloat{listing}{tb}{lst}{}
\floatname{listing}{Listing}
\usepackage[frozencache,cachedir=.]{minted}
\usepackage{tcolorbox}
\usepackage{enumitem}
\usepackage{booktabs}

\newenvironment{tightlist}%
{\begin{list}{$\bullet$}{%
    \setlength{\topsep}{0in}
    \setlength{\partopsep}{0in}
    \setlength{\itemsep}{0in}
    \setlength{\parsep}{0in}
    \setlength{\leftmargin}{1.5em}
    \setlength{\rightmargin}{0in}
}
}%
{\end{list}
}

\title{Generalized Planning in PDDL Domains with Pretrained Large Language Models}
\author{
    Tom Silver\textsuperscript{\rm 1},
    Soham Dan\textsuperscript{\rm 2},
    Kavitha Srinivas\textsuperscript{\rm 2},\\
    Joshua Tenenbaum\textsuperscript{\rm 1},
    Leslie Kaelbling\textsuperscript{\rm 1},
    Michael Katz\textsuperscript{\rm 2}
}
\affiliations{
    \textsuperscript{\rm 1}MIT Computer Science and Artificial Intelligence Laboratory;
    \textsuperscript{\rm 2}IBM Research\\
    Correspondence: tslvr@mit.edu, Michael.Katz1@ibm.com
}

\usepackage{bibentry}

\begin{document}

\maketitle

\begin{abstract}
Recent work has considered whether large language models (LLMs) can function as planners: given a task, generate a plan.
We investigate whether LLMs can serve as generalized planners: given a domain and training tasks, generate a program that efficiently produces plans for other tasks in the domain.
In particular, we consider PDDL domains and use GPT-4 to synthesize Python programs.
We also consider (1) Chain-of-Thought (CoT) summarization, where the LLM is prompted to summarize the domain and propose a strategy in words before synthesizing the program; and (2) automated debugging, where the program is validated with respect to the training tasks, and in case of errors, the LLM is re-prompted with four types of feedback.
We evaluate this approach in seven PDDL domains and compare it to four ablations and four baselines.
Overall, we find that GPT-4 is a surprisingly powerful generalized planner.
We also conclude that automated debugging is very important, that CoT summarization has non-uniform impact, that GPT-4 is far superior to GPT-3.5, and that just two training tasks are often sufficient for strong generalization.
\footnote{Code and logs: \url{https://github.com/tomsilver/llm-genplan/}}
\end{abstract}

\section{Introduction}

While some classes of sequential decision-making tasks are provably intractable \cite{chapman-aij1987}, others can be solved efficiently with a single domain-specific program.
In the latter case, there is considerable interest in \emph{automatically synthesizing} these programs given a small number of training tasks.
In AI planning, several approaches to this \emph{generalized planning} problem have been proposed, with programs expressed as lifted decision lists, as finite state machines, or in domain-specific languages~\cite{srivastava2011foundations,bonet2015policies,jimenez2019review,rivlin2020generalized}.
In reinforcement learning, goal-conditioned policies and value functions can be understood as particular kinds of programs learned with the same generalized planning objective~\cite{gcrl1,gcrl2}.
Despite these efforts, it remains challenging to efficiently synthesize programs from few training tasks that generalize to a wide variety of held-out tasks. 

Given the tremendous recent progress in large language models (LLMs)~\citep{brown2020language,chen2021evaluating,chowdhery2022palm}, especially in code generation~\cite{chen2021evaluating,nijkamp2023codegen,chen2023improving}, this work asks a simple question: \emph{can pretrained LLMs be used for generalized planning?}
In particular, we investigate whether GPT-4~\cite{openai2023gpt4} can be used to write a domain-specific Python program that solves a set of tasks in a planning domain.
For each domain, we prompt GPT-4 with the domain and a small number of training tasks encoded in the Planning Domain Definition Language (PDDL)~\cite{mcdermott-aimag2000}.
We then ask GPT-4 to write a program that consumes a (parsed) task description and outputs a plan.
To prevent it from writing domain-general search-based code---a natural inclination given the association between PDDL and search in its pretraining data---we instruct GPT-4 to implement ``a simple strategy that does not use search.''

\begin{figure*}[t]
    \centering
    \includegraphics[width=0.8\textwidth]{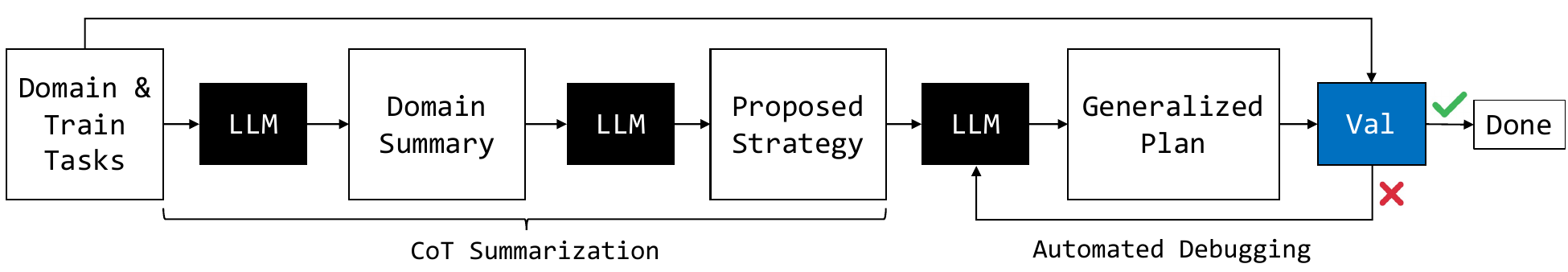}
    \caption{Overview of pipeline for generalized planning with pretrained LLMs. See text for details.}
    \label{fig:teaser}
\end{figure*}

Beyond this basic protocol, we consider two extensions.
First, inspired by Chain-of-Thought (CoT)~\cite{wei2022chain,jiang2023self}, we prompt GPT-4 to write a natural language \emph{summary} of the PDDL domain.
We then ask it to describe a solution strategy before finally implementing the strategy in Python.
Second, inspired by Inner Monologue~\cite{huang2022inner} and Corrective Re-prompting~\cite{raman2022planning}, we automatically provide feedback to GPT-4 in the case where it fails to solve training tasks.
For example, if executing the Python code results in an exception, we present GPT-4 with that exception and ask it to fix the code.
We repeat this \emph{automated debugging} process up to four times or until all training tasks are solved.
See Figure~\ref{fig:teaser} for an overview of this pipeline.

In our experiments, we evaluate this approach on seven PDDL domains: six from recent work in generalized planning~\cite{yang2022pg3}, and a seventh novel domain.
We find that the approach is a strong baseline compared to existing generalized planning approaches.
This is an important finding that we expect to inform further research in generalized planning.
We also present a suite of ablations and additional analyses to unpack the contributions of CoT summarization, automated debugging, names in the PDDL, and GPT-4 vs. GPT-3.5.
Our results suggest that automated debugging, PDDL names, and GPT-4 are very important, while the impact of CoT is non-uniform.
Finally, we provide qualitative analyses of common failure cases, suggesting directions for future work.
We conclude that GPT-4 is a surprisingly powerful generalized planner when properly guided.

\section{Related Work}

\textbf{LLMs for (PDDL) Planning.}
\emph{Generalized planning} with LLMs can be seen as an alternative to \emph{planning} with LLMs~\cite{sharma2022skill,ahn2022can,huang2022language,raman2022planning,lin2022grounded}.
Most relevant is work by~\citet{valmeekam2022large,silver2022pddl} who consider LLM-based planning in PDDL domains.
There are several advantages to using LLMs for generalized planning, rather than planning: (1) programs produced by the LLM can be inspected and validated; (2) running a synthesized program can be much faster (and cheaper) than querying the LLM for each new task; (3) synthesized programs can scale to arbitrarily large tasks, whereas current LLMs are limited by context window size.
\citet{pallagani2022plansformer} consider fine-tuning an LLM to solve PDDL tasks.
Other recent work has considered using LLMs for translating between natural language and PDDL~\cite{collins2022structured,lin2023text2motion,xie2023translating,liu2023llm}.
These efforts could be combined with our approach.

\textbf{Generalized Planning.}
This work contributes to a growing literature on generalized planning~\cite{triangletables,jimenez2019review}.
Prior work has considered synthesizing generalized plans in several ways: (1) performing a search through a hypothesis class of generalized policies~\cite{levine2003learning,jimenez2015computing,segovia2018computing,segovia2021generalized}; (2) using example plans to construct a generalized plan, often represented with a finite-state machine~\cite{kplanner,Srivastava2011DirectedSF,winner2008dsplanner}; and (3) discovering state and action abstractions and then using them in a generalized plan ~\cite{genplan_representation}.
One pervasive challenge is that there are often many valid plans for any given task, and only some of these plans are consistent with a simple generalized plan.
PG3 addresses this challenge by using candidate generalized plans (represented as lifted decision list goal-conditioned policies) to constrain the generation of example plans~\cite{yang2022pg3}.
We use PG3 as the main point of comparison in experiments.

\textbf{LLMs for Code Generation.}
Our work builds on recent techniques that use LLMs for code generation~\cite{chen2021evaluating,nijkamp2023codegen}.
CoT summarization is related to several techniques that ask the LLM to outline its ``thinking'' before arriving at a final implementation~\cite{wei2022chain,jiang2023self,zheng2023outline}.
A number of recent works also use programs as prompts (i.e.,  a structured chain of thought) in an attempt to help LLMs perform mathematical reasoning~\cite{gao2022pal,imani2023mathprompter}.  
Related to our automated debugging, \citet{xia2023conversational,chen2023teaching} consider automated program repair by re-prompting the LLM with feedback from failed validation checks.
\citet{chen2023improving} consider a related paradigm, but where feedback comes from humans, rather than automated checks.
Also relevant are efforts to generate code that can be used for robotic decision-making~\cite{liang2022code,singh2022progprompt}.
Beyond LLMs, code generation has been studied extensively in program synthesis~\cite{alur2013syntax,gulwani2017program} and inductive logic programming~\cite{muggleton1991inductive,cropper2022inductive}.

\section{Background and Problem Setting}

\textbf{PDDL Domains and Tasks.}
We consider deterministic, fully-observed planning tasks represented in PDDL.
In experiments, we use the STRIPS subset with types and negative preconditions.
We describe PDDL informally and refer the reader to other references for a formal treatment~\cite{mcdermott-aimag2000}.
A PDDL \emph{domain} is characterized by a name, a set of types, a set of predicates, and a set of operators.
For example, in the Delivery domain, a robot must pick up newspapers from a home base and then deliver them to certain locations.
The domain has two types: \texttt{loc} and \texttt{paper}.
One predicate is \texttt{(at ?l - loc)}, where \texttt{?l} is a placeholder for a \texttt{loc} object.
The domain has three operators: \texttt{(pick-up ?p - paper ?l - loc)}, \texttt{(move ?from - loc ?to - loc)}, \texttt{(deliver ?p - paper ?l - loc)}.
For example, the \texttt{pick-up} operator in its entirety is:
{\footnotesize
\begin{verbatim}
(:action pick-up
  :parameters (?p - paper ?l - loc)
  :precondition (and (at ?l)
    (isHomeBase ?l)
    (unpacked ?p))
  :effect (and
    (not (unpacked ?p))
    (carrying ?p)))
\end{verbatim}
}

A PDDL \emph{task} is characterized by a domain, a set of objects, an initial state, and a goal.
An object has a name and a type, e.g., \texttt{paper1 - paper}.
A ground atom is a predicate and a tuple of objects of the appropriate types, e.g., \texttt{(unpacked paper1)}.
A state consists of a conjunction of ground atoms that are true, assuming all other ground atoms to be false.
A goal is a conjunction of ground atoms that must be true in any {\em goal state}.
(More general goal expressions are also possible in PDDL.)
For example, in Delivery, the goal may include \texttt{(satisfied loc1)} and \texttt{(satisfied loc2)}.

An \emph{action} is an operator and a tuple of objects of the appropriate types, e.g., \texttt{(pick-up paper1 loc4)}.
The operator's preconditions determine whether the action is applicable and the effects define what ground atoms would be added or deleted if the operator is executed.
A \emph{plan} is a finite sequence of actions.
The plan is \emph{valid} for a task if all actions are applicable when executed in succession from the initial state and if the final state is a goal state.

PDDL domains, types, predicates, operators, objects, and types often include human-readable names like the ones shown above.
These names are not important for standard AI planners or previous generalized planning approaches.
However, the names are very important for humans---and, we expect, for LLMs---trying to make sense of the PDDL.

\textbf{Generalized Planning in PDDL Domains.}
A generalized planning instance is characterized by a PDDL domain and a distribution of tasks.
A small set of \emph{training tasks} (10 or fewer in experiments) from the distribution is given at training time.
A set of held-out \emph{evaluation tasks}---typically involving many more objects---are used to measure performance.
The objective is to use the training tasks to synthesize a program that will produce valid plans for all of the evaluation tasks.
We consider an evaluation task solved if the program returns a valid plan within a fixed wall-clock time budget (30 seconds in experiments).
In other words, we are interested in satisficing, not optimal, planning, and our primary concern is the efficiency of planning itself.

\section{Generalized Planning with LLMs}

We are interested in the extent to which pretrained large language models (LLMs) can be used for generalized planning in PDDL domains.
We assume familiarity with LLMs~\citep{brown2020language,chen2021evaluating,chowdhery2022palm,openai2023gpt4}.
To use LLMs for generalized planning, we need to define a protocol for prompting.

\subsection{Prompting Protocol}

Previous work on Chain-of-Thought (CoT) prompting has shown that asking an LLM to ``think step by step'' can improve performance in reasoning tasks~\cite{wei2022chain}.
With these results in mind, we hypothesized that decomposing generalized planning into three stages---domain summarization, strategy proposal, and strategy implementation---would improve performance.

\textbf{Domain Summarization.}
Our first prompt to the LLM is in the following form:

\begin{tcolorbox}[left=2pt,right=2pt]
Domain: [PDDL Domain]

Example problems: [PDDL Training Tasks]

Write a short summary of this domain in words.
\end{tcolorbox}

To compensate for the limited context window size of transformer-based LLMs like GPT-4, we abbreviate the encoding of the training tasks in two ways.
First, we always use only two training tasks, even when more are given.
Second, within each training task, we limit the number of objects and initial state ground atoms shown.
For each object type, if the number of objects of that type exceeds 10, we truncate the object set and add ellipses.
Similarly, for each predicate, if the number of ground atoms with that predicate exceeds 10, we truncate and add ellipses.
The fact that we only need to communicate the ``gist'' of the task distribution, rather than whole tasks, is another advantage of generalized planning with LLMs versus \emph{planning} with LLMs.

\textbf{Strategy Proposal.}
After the LLM responds to the first prompt, we ask for a generalized planning strategy:

\begin{tcolorbox}[left=2pt,right=2pt]
There is a simple strategy for solving all problems in this domain without using search. What is that strategy?
\end{tcolorbox}

In preliminary experiments, omitting the phrase ``without using search'' would often lead the LLM to propose a search-based planning strategy.

\textbf{Strategy Implementation.}
Finally, we ask the LLM to implement the strategy as a Python program:

\begin{tcolorbox}[left=2pt,right=2pt]
\footnotesize
Implement the strategy as a Python function. The code should be of the form

\begin{verbatim}
def get_plan(objects, init, goal):
    # Your code here
    return plan    
\end{verbatim}

where
\begin{itemize}
    \item \texttt{objects} is a set of (object name, type name) tuples
    \item \texttt{init} is a set of ground atoms represented as tuples of predicate names and arguments (e.g., (`predicate-foo', `object-bar', ...))
    \item \texttt{goal} is also a set of ground atoms represented in the same way
    \item \texttt{plan} is a list of actions, where each action is a ground operator represented as a string (e.g., `(operator-baz object-qux ...)')
\end{itemize}
\end{tcolorbox}

In domains without object types, \texttt{objects} is instead just a set of object names.

\subsection{Automated Interactive Debugging}

After the LLM has proposed an implementation of \texttt{get\_plan}, we use the training tasks to \emph{validate} the implementation.
For each training task, we execute \texttt{get\_plan} until it returns an output, throws an exception, or reaches a timeout (30 seconds).
If the output is a valid plan, we continue onto the next training task.
Otherwise, we re-prompt with one of four types of feedback.

\textbf{Python Exceptions.} If executing \texttt{get\_plan} results in a Python exception, we capture the traceback and report it to the LLM along with the input.
An example is shown below, with the traceback abbreviated for clarity.
\begin{tcolorbox}[left=2pt,right=2pt]
Given this task: [PDDL Training Task]

The code raised the following exception:
\begin{verbatim}
File "<file-name-omitted>", line 86
lift_at = {atom[1]: atom[2] ...}
                    ~~~~^^^
IndexError: tuple index out of range
\end{verbatim}

Fix the code.
\end{tcolorbox}

In preliminary experiments, we found that including the full traceback can improve performance.

\textbf{Timeout.} If \texttt{get\_plan} does not finish before the timeout, we report to the LLM that the program did not finish and suggest that an infinite loop may be to blame.
We also provide a traceback showing where the program was executing when it was interrupted.
An example is shown below.

\begin{tcolorbox}[left=2pt,right=2pt]
Given this task: [PDDL Training Task]

The code raised the following exception:
\begin{verbatim}
File "<file-name-omitted>", line 23
while not any(span_loc[1] == ...:
^^^^^^^^^^^^^^^^^^^^^^^^^^^^^^^^^
KeyboardInterrupt
\end{verbatim}
The code was interrupted because it timed out (possible infinite loop).

Fix the code.
\end{tcolorbox}

The traceback is again abbreviated for clarity.
Note that the \texttt{KeyboardInterrupt} is automatically thrown after 30 seconds.
In practice, nearly all timeouts we observe are due to logic errors in the code, rather than inefficient but correct implementations.

\textbf{Plan Syntax.} If \texttt{get\_plan} returns an output, we check its syntax: whether it is a list of strings, whether each string is enclosed in parentheses and space-separated, and whether the action names, object names, and number of objects per action are valid with respect to the domain and task.
If any of these checks fail, we report the failure to the LLM.
For this type of failure, we also remind the LLM about the valid operators.
An example is shown below.

\begin{tcolorbox}[left=2pt,right=2pt]
Given this task: [PDDL Training Task]

The code returned this plan: 
\begin{verbatim}
['walk r0_c0 r0_c1', 'walk ...]
\end{verbatim}
However, the action \texttt{walk r0\_c0 r0\_c1} is invalid at step 0. NOTE: the valid operators are: \texttt{(climb ?from ?to) (walk ?from ?to)}.

Fix the code.
\end{tcolorbox}

The full plan is shown to the LLM but abbreviated in the example for clarity.
The issue in this example is that the actions are not enclosed in parentheses.

\textbf{Plan Semantics.}
If all of the previous checks pass, we use the VAL tool~\cite{howey2004val} to check whether the \texttt{get\_plan} output is a semantically valid plan.
If not, VAL provides ``plan repair advice'', e.g., if there is an action with invalid preconditions.
We extract this plan repair advice and report it to the LLM.
Note that we use this advice not to repair the \textit{plan}, but rather, to repair the \emph{generalized plan}.
An example is shown below.

\begin{tcolorbox}[left=2pt,right=2pt]
Given this task: [PDDL Training Task]

The code failed. It returned the following plan:
\begin{verbatim}
['(pick-up paper-1 loc-0)', ...].
\end{verbatim}
NOTE: \texttt{(pick-up paper-0 loc-0)} has an unsatisfied precondition at time 3
\begin{verbatim}
(Set (at loc-0) to true)
\end{verbatim}
Fix the code.
\end{tcolorbox}

\textbf{Additional Details.}
After re-prompting the LLM, we repeat the process of checking the code and reporting any failures up to four times.
To handle rare cases where the LLM implements its own helper functions and then assumes during debugging that the helper functions are still available, we append each new response from the LLM to a growing Python file, rather than overwriting the previous responses.
If a failure is still encountered on the last attempt, the final response is used during evaluation.

\section{Experiments and Results}

\begin{table*}[t]
\centering
\begin{tabular}{cccccccccc} 
 \toprule
 \textbf{Domain} & GPT-4 & No CoT & No Debug & No Names & GPT-3.5 & PG3 & Policy Eval & Plan Compare & Random \\
 \midrule
 Delivery & 0.90 & 0.70 & 0.10 & 0.10 & 0.00 & 1.00 & 0.00 & 0.10 & 0.00 \\ 
 Forest & 1.00 & 1.00 & 0.62 & 0.11 & 0.32 & 1.00 & 1.00 & 0.16 & 0.03 \\ 
 Gripper & 0.90 & 0.80 & 0.50 & 0.10 & 0.00 & 1.00 & 0.00 & 0.20 & 0.00 \\ 
 Miconic & 0.01 & 0.13 & 0.00 & 0.00 & 0.00 & 1.00 & 0.00 & 0.10 & 0.13 \\ 
 Ferry & 0.80 & 0.20 & 0.26 & 0.00 & 0.00 & 1.00 & 0.00 & 0.90 & 0.00 \\ 
 Spanner & 0.10 & 0.00 & 0.00& 0.00 & 0.00  & 1.00 & 1.00 & 0.56 & 0.06 \\ 
 Heavy & 0.60 & 1.00 & 0.20 & 0.00 & 0.00 & 0.00 & 0.00 & 0.00 & 0.00 \\ 
 \bottomrule
\end{tabular}
\caption{Fraction of evaluation tasks solved. All results are averaged over 10 random seeds and 30 evaluation tasks per seed.}
\label{table:main_results}
\end{table*}

\begin{figure*}[t]
    \includegraphics[width=1.0\textwidth,trim={0 0.5cm 0 0},clip]{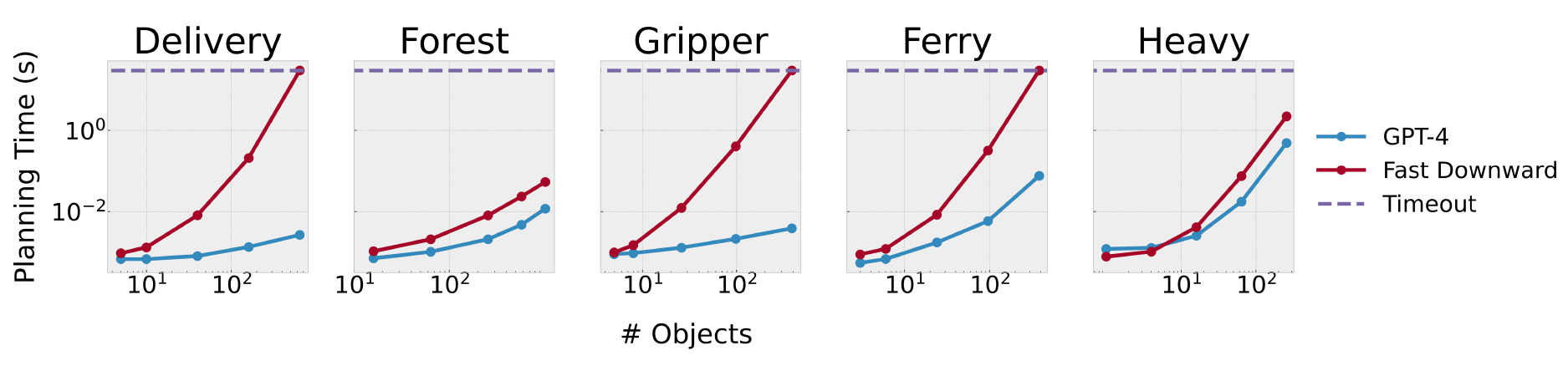}
    \caption{GPT-4 synthesized program runtime compared to a state-of-the-art planner (Fast Downward). Note the log-log axes.
    Each point is a median over 10 newly generated tasks, over all seeds where generalized planning solved all evaluation tasks.}
    \label{fig:problem_size}
\end{figure*}

\begin{figure}[t]
    \centering
    \includegraphics[width=0.6\columnwidth]{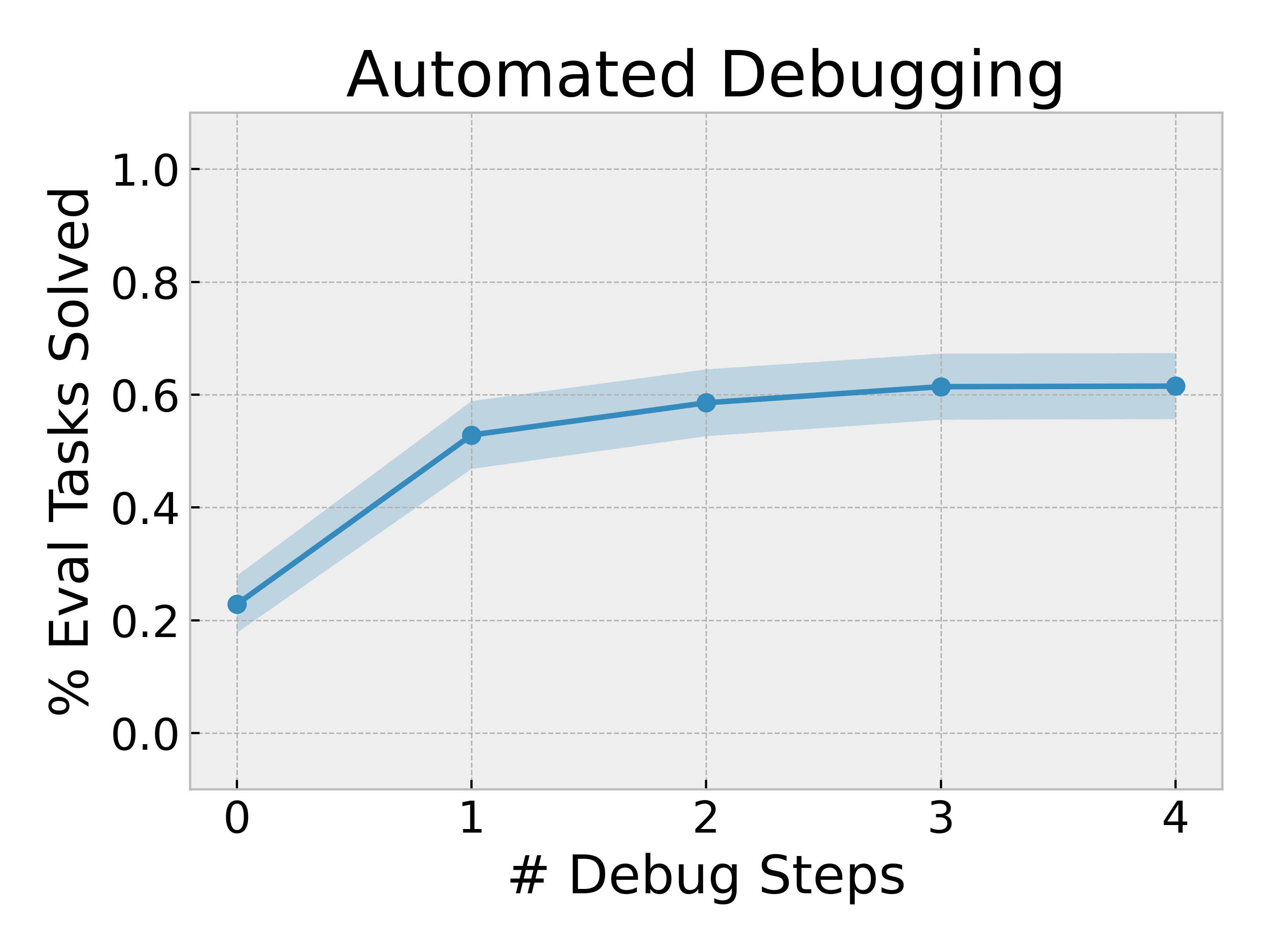}
    \caption{Fraction of evaluation tasks solved by GPT-4 versus number of debugging steps allowed, averaged over all domains and seeds. The shaded region is standard error.}
    \label{fig:automated_debugging}
\end{figure}

\begin{table}[t]
\centering
\begin{tabular}{lccc} 
 \toprule
 \textbf{Error Type} & All & Success & Failure \\
 \midrule
 Python Exception & 40.0 & 28.9 & 42.5 \\
 Plan Semantics & 34.0 & 44.7 & 31.4 \\
 Plan Syntax & 13.0 & 18.4 & 11.7 \\
 Timeout & 13.0 & 8.0 & 14.4 \\
 \bottomrule
\end{tabular}
\caption{Percentages of error types encountered by GPT-4 in training tasks over all domains and seeds. ``All'' is the breakdown for all training tasks; ``Success'' is the breakdown for trials where all evaluation tasks were subsequently solved; ``Failure'' is the breakdown for the non-Success trials.}
\label{table:error_types}
\end{table}

Through experiments, we address these questions:
\textbf{1.} Can GPT-4 be used for generalized (PDDL) planning?
\textbf{2.} Are the synthesized programs efficient?
\textbf{3.} Does CoT summarization help?
\textbf{4.} Does automated debugging help?
\textbf{5.} To what extent does GPT-4 rely on names in the PDDL?
\textbf{6.} How does GPT-4 compare to GPT-3.5?
\textbf{7.} Do each of the four error types help?
\textbf{8.} How many training tasks are needed?

\subsection{Experimental Setup}
We evaluate nine generalized planning approaches on seven PDDL domains over 10 random seeds.
Tasks are randomly generated for each seed.

\textbf{Domains.}
The first six domains (and tasks) are taken directly from the previous work by~\citet{yang2022pg3}.
Of these, four (Gripper, Miconic, Ferry, Spanner) are standard planning benchmarks and the other two (Delivery, Forest) were introduced by that work.
The last domain (Heavy) is new to this work.
The pretraining data for GPT-4 is not publicly available, but it is likely that the \emph{domain definitions} for at least the four standard domains were included in that data.
However, we believe it is unlikely that \emph{generalized plans} were included, and for the Heavy domain, we can guarantee that neither the domain nor generalized plans were included.
We now briefly describe each domain.
Unless otherwise specified, there are 10 training tasks and 30 evaluation tasks per domain and seed.
\vspace{2pt}
\begin{tightlist}
\item \textbf{Delivery}: Newspapers at a home base must be delivered to multiple locations. There are five training tasks with 9--17 objects; evaluation tasks have 70--100 objects.
\item \textbf{Forest}: A hiker must navigate a 2D grid to reach a goal location while climbing hills and avoiding water. A marked trail leads to the goal, but there are shorter paths through dirt. There are 4 training tasks with 64-100 objects; evaluation tasks have 100--144 objects.
\item \textbf{Gripper}: Balls must be transported between rooms by a robot with two grippers. Training tasks have 20--30 objects; evaluation tasks have 60--80 objects.
\item \textbf{Miconic}: Passengers in multiple buildings, each with an elevator, must be picked up and dropped off on different floors. Training tasks have 6--30 objects; evaluation tasks have 11--150 objects.
\item \textbf{Ferry}: Cars must be sailed between islands using a ferry that can carry at most one car.
Training tasks have 13--20 objects; evaluation tasks have 30--50 objects.
\item \textbf{Spanner}: Wrenches (spanners) and nuts are distributed along a one-way corridor. An agent must move down the corridor, pick up wrenches, and tighten the nuts, using each wrench at most once.
Training tasks have 9--15 objects; evaluation tasks have 30--60 objects.
\item \textbf{Heavy}: Items must be stacked into an empty box. An item can only be stacked on another item if the latter is heavier. The weight relations are expressed via a \texttt{(heavier ?x ?y)} predicate. One challenge is in determining which item to place into the box first, i.e., which item is the heaviest.
Training tasks have 3--10 objects; evaluation tasks have 100--250 objects.
\end{tightlist}
\vspace{2pt}

\textbf{Approaches.}
We evaluate the main approach, four ablations, and four baselines.
The baselines are taken from the work by~\citet{yang2022pg3}; see that work for details.

\vspace{2pt}
\begin{tightlist}
\item \textbf{GPT-4}: Our main approach with CoT summarization and automated debugging.
\item \textbf{No CoT}: An ablation of the main approach that does not use CoT summarization. The three initial prompts are combined and ``Write a short summary of this domain in words.'' and ``What is that strategy?'' are removed.
\item \textbf{No Debug}: An ablation of the main approach that does not use automated debugging. The first implementation of \texttt{get\_plan} is used for evaluation.
\item \textbf{No Names}: An ablation of the main approach where all names in the PDDL domains and tasks are replaced with nondescriptive identifiers. For instance, predicates are renamed to \texttt{predicate1}, \texttt{predicate2}, etc., operators are renamed to \texttt{operator1}, \texttt{operator2}, etc. Altogether, the names of the domain, problem, predicates, operators, variables, types, and objects are ablated.
\item \textbf{GPT-3.5}: GPT-3.5 with CoT summarization and automated debugging.
\item \textbf{PG3}: The generalized planning approach proposed by \citet{yang2022pg3}. The synthesized programs are goal-conditioned policies implemented as lifted decision lists. Synthesis is performed via heuristic search in policy space with their novel heuristic.
\item \textbf{Policy Evalulation (PE)}: An approach from \citet{yang2022pg3} that is identical to PG3 except that the heuristic used for policy search is sparse: each candidate policy is scored based on the number of training tasks solved.
\item \textbf{Plan Compare (PC)}: Another approach from \citet{yang2022pg3} that is identical to PG3 except for the policy search heuristic: example plans for each training task are generated offline, and the policy is scored based on its agreement with the example plans.
\item \textbf{Random}: Valid actions are randomly sampled and executed until a dead-end is encountered, the goal is reached, or a maximum horizon (default 1000, but see the previous work) is exceeded.
\end{tightlist}
\vspace{2pt}

\textbf{Experimental Details.}
We used a Macbook Pro laptop with an M1 chip and 64 GB RAM.
Since an API for GPT-4 is not publicly available, we used the ChatGPT browser interface for all experiments (including the GPT-3.5 baseline).
The pipeline is fully automated except that prompts and responses are manually copied and pasted between the terminal and browser, with the clipboard programmatically updated.
To facilitate reproducibility, we have released all chat logs and code.

\subsection{Results and Analysis}

Main results are presented in Table~\ref{table:main_results}.
Examples of synthesized programs are presented in the appendix.
Overall, the performance of GPT-4 with CoT summarization and automated debugging is strong in Delivery, Forest, Gripper, Ferry, and Heavy, and poor in Miconic and Spanner.
Note that the reported success rates are averaged over all LLM conversations.
In practice, performance could be boosted by restarting the conversation multiple times and using the best-found program~\cite{chowdhery2022palm}.
The strong performance in Heavy is especially notable.
The generalized planning baselines fail in this domain because lifted decision lists are overly restrictive as program representations and cannot discover a concept like ``heaviest overall'' from pairwise heavier relations.
GPT-4's ability to write general Python code is one of its biggest advantages as a generalized planning approach.

We also observe that in nearly all cases, GPT-4 either (1) solves all of the training tasks and then solves all of the evaluation tasks; or (2) fails to solve at least one training task and then fails to solve all of the evaluation tasks.
In other words, overfitting to the training tasks is very rare, and evaluation performance is typically all-or-nothing.
See Table~\ref{table:max_results} in the appendix for the \emph{maximum} fraction of tasks solved.

\textbf{Miconic failures.}
GPT-4 has a number of consistent failure modes in Miconic.
First, at the strategy proposal level, it often fails to recognize that there can be multiple buildings, each with their own elevator.
This is admittedly difficult to recognize given the PDDL encoding: buildings exist only implicitly based on the \texttt{above} relation between floor objects.
For example, one would need to see that neither \texttt{(above f1\_b1, f1\_b2)} nor \texttt{(above f1\_b2, f1\_b1)} are true and conclude that the floors are in two different buildings.
However, especially after automated debugging, GPT-4 \emph{can} realize that there are multiple buildings, and furthermore, that building names (e.g., \texttt{b1}, \texttt{b2}) can be extracted from the floor names.
But then other failures often occur, for example, attempting and failing to create a total ordering of the floors from the \texttt{above} predicate.
Overall, we believe that Miconic is just beyond the limit of GPT-4's current capabilities and would likely be solved by the next generation of LLMs, or by GPT-4 with additional guidance.

\textbf{Spanner failures.}
GPT-4 consistently fails in Spanner during strategy proposal.
In particular, GPT-4 does not appear to realize that locations in Spanner are connected in a \emph{one-way} chain.
The strategy proposed is often ``first collect all of the spanners, then tighten all of the nuts'' or similar.
A correct strategy would instead be to ``move to each location in the chain, picking up any spanners and tightening any nuts at each location.''
Recognizing the existence of the one-way chain requires examining the \texttt{link} atoms in the training problems.
Even after automated debugging, GPT-4 often assumes, incorrectly, that links are commutative.

\textbf{Program efficiency.}
Although we prompt the LLM to implement a ``simple'' program that does not use search, it is still possible for the LLM to produce a program that does use search or is slow for other reasons (e.g., poor algorithmic complexity).
We therefore measure synthesized program runtime. As a baseline for our comparison we use a state-of-the-art domain-independent PDDL planner LAMA \cite{richter2010lama} via Fast Downward~\cite{fd}, stopping after the first plan is found.\footnote{Our intention is not to compare planners, but rather to provide a frame of reference for runtime.}
In Figure~\ref{fig:problem_size}, we plot wall-clock runtimes as a function of problem size (number of objects).
Overall, we see that the synthesized programs not only scale favorably with respect to the planner, but also consistently beat the planner in absolute runtime by large margins.
This is notable given that the LLM synthesizes Python programs, while the PDDL planner uses a highly optimized combination of Python and C++ code.
The bottleneck for Fast Downward is often operator grounding.
The LLM's programs do not need to ground operators---they can go directly from task to plan.

\textbf{The role of CoT.}
Comparing GPT-4 to No CoT, we see that the impact of CoT summarization is mixed: it seems to help in most cases, but hurt in Miconic and Heavy.
Miconic is an especially interesting case.
When using CoT summarization, GPT-4 nearly always proposes a ``sweep'' strategy, where the elevator(s) are first moved to the bottom floor; then moved up one floor at a time until the top floor, picking up and dropping off passengers along the way; then moved down one floor at a time, again picking up and dropping off passengers.
This strategy would work in theory, but it requires finding a total ordering of floors within buildings.
Without CoT, GPT-4 often attempts a different strategy: pick up, move, and drop off each passenger, one at a time.
The latter strategy does not require a total ordering over floors and is arguably simpler to implement in Python.
This example shows that CoT can influence the strategy proposed by GPT-4.
Moreover, strategies that are ``simple'' to describe in natural language may not be simple to implement in code.
In Heavy, there is not a clear difference in strategies with and without CoT.
Since a good strategy is evidently discernible from the PDDL alone, it is possible that CoT ``distracts'' GPT-4 during implementation.

\textbf{The role of automated debugging.}
Comparing GPT-4 to No Debug, we see that automated debugging generally improves performance dramatically.
Figure~\ref{fig:automated_debugging} shows that even one step of automated debugging helps substantially, and further steps exhibit diminishing marginal improvements.
Table~\ref{table:error_types} reports the fraction of error types encountered during training across.
Python exceptions are most common, followed closely by errors in plan semantics, then errors in plan syntax, and finally timeouts.
We also see that the error types are well-distributed within successful trials, suggesting that each of the four types of feedback are beneficial.
In general, GPT-4 tends to make small, local corrections to code during automated debugging.
If the code is structurally flawed and requires a significant rewrite, restarting the dialogue from the beginning may be required.

\textbf{The role of PDDL names.}
Examining the results for the No Names ablation, we see performance overall is very poor.
This confirms our hypothesis that the terms present in the PDDL domains and tasks are helpful to the LLM, as they would be to a human.
Note that planners like Fast Downward and generalized planners like PG3 would be unaffected by name changes.
However, there are a few cases where the No Names ablation does succeed, suggesting that the LLM has some capacity for purely syntactic generalized planning.

\textbf{GPT-3.5 vs. GPT-4.}
Examining the results for GPT-3.5, we see that it performs much worse than GPT-4.
This is consistent with other reports~\cite{openai2023gpt4,bubeck2023sparks} that GPT-4 is far superior on reasoning and coding tasks.
Qualitatively, the programs proposed by GPT-3.5 are flawed in myriad ways and do not usually appear ``close''.
They also do not seem to improve with automated debugging.

\textbf{Data efficiency.}
In the appendix, we analyze the number of training tasks \emph{used} in each successful trial.
A training task is used if it appeared in the prompt and/or triggered feedback during automated debugging.
Since two training tasks are always used in the prompt, the minimum used is two.
Interestingly, in the vast majority of cases, only those two training tasks are used.
During automated debugging, these two prompting tasks are always checked first, and most of the time, they are sufficient to identify issues.
In a small number of cases, a third task is also used during automated debugging.
This result speaks to the strong few-shot learning capabilities of GPT-4.
We expect that in many cases, even one training task would suffice, although we did witness a drop in performance in preliminary experiments with one task.

\section{Discussion and Future Work}

In this work, we showed that GPT-4 with CoT summarization and automated debugging is a surprisingly strong generalized planner in PDDL domains.
We conclude with limitations of this work, reflections about the implications of our findings, and opportunities for future work.

\textbf{Limitations.}
A major limitation of this work and previous work on generalized planning is that it is easy enough to hand-design generalized plans for all of the domains considered.
Nonetheless, we expect this line of work to be practically useful for at least three reasons.
(1) In some cases, it may be considerably easier to specify PDDL domain and problem descriptions than it is to directly specify a generalized plan.
(2) In a fully autonomous system, where operators and predicates are learned \emph{in association to natural language}, we would want the system to also synthesize generalized plans autonomously.
(3) Beyond PDDL, generalized planning with LLMs would be an even more attractive option, since other approaches rely strongly on formal specifications.
Another limitation of this work is our use of training tasks to communicate the task distribution of interest to the LLM.
In general, a few example tasks may be insufficient to express the full distribution.
Other representations like natural language or procedural generation code may be better, but would require more human input.

\textbf{Is (generalized) planning now obsolete?}
No.
First, there remains a performance gap between GPT-4 and PG3, and other generalized planners may be even better.
However, even if this gap is closed by the next generation of LLMs, we would still say no.
\emph{Planning} remains essential in domains where no simple program exists.
An interesting direction for future work would be automatically detecting whether a simple program might exist before attempting to synthesize one.
We tried the Sokoban domain and found that GPT-4 correctly indicates that no simple program exists.
However, this property of Sokoban is well-known, so it is likely parroting pretraining data.
We also tried the Slitherlink domain, which was featured in the 2023 International Planning Competition, and found that GPT-4 did \emph{not} recognize that no simple strategy exists~\cite{Yato2003OnTN}.
\emph{Generalized} planning without LLMs also remains important in cases where domain descriptions are not human-readable, e.g., because the predicates or operators are learned 
\cite{silver2023inventing}.
Even with natural language descriptions, combining ``classical'' approaches with LLMs may be best.

\textbf{What if we gave the LLM access to a planner?}
Giving an LLM access to APIs is a very powerful idea~\cite{schick2023toolformer} and one such API could be a PDDL planner~\cite{liu2023llm}.
An LLM could potentially use such a planner for generalized planning, especially given that approaches like PG3 rely on access to a planner to generate example plans.
In some domains, generating example plans naively would likely confuse the LLM.
For example, plans generated in the Forest domain would follow arbitrary paths through the dirt rather than following the slightly longer marked trail.
In other cases, though, example plans could be very useful, especially if the LLM generates them in a targeted way.
Leveraging \emph{diverse} plans~\cite{sohrabi2016finding,katz2020reshaping} could be particularly useful.

\clearpage

\bibliography{aaai24}

\clearpage

\appendix

\section{Additional Results}

Table~\ref{table:max_results} reports the maximum fraction of evaluation tasks solved over seeds.
In most cases, the LLM either solves all or none of the evaluation tasks.
This suggests that the LLM does not overfit to the training tasks, even though only a very small number of them are used (see main text).

\begin{figure}[h]
    \centering
    \includegraphics[width=0.8\columnwidth]{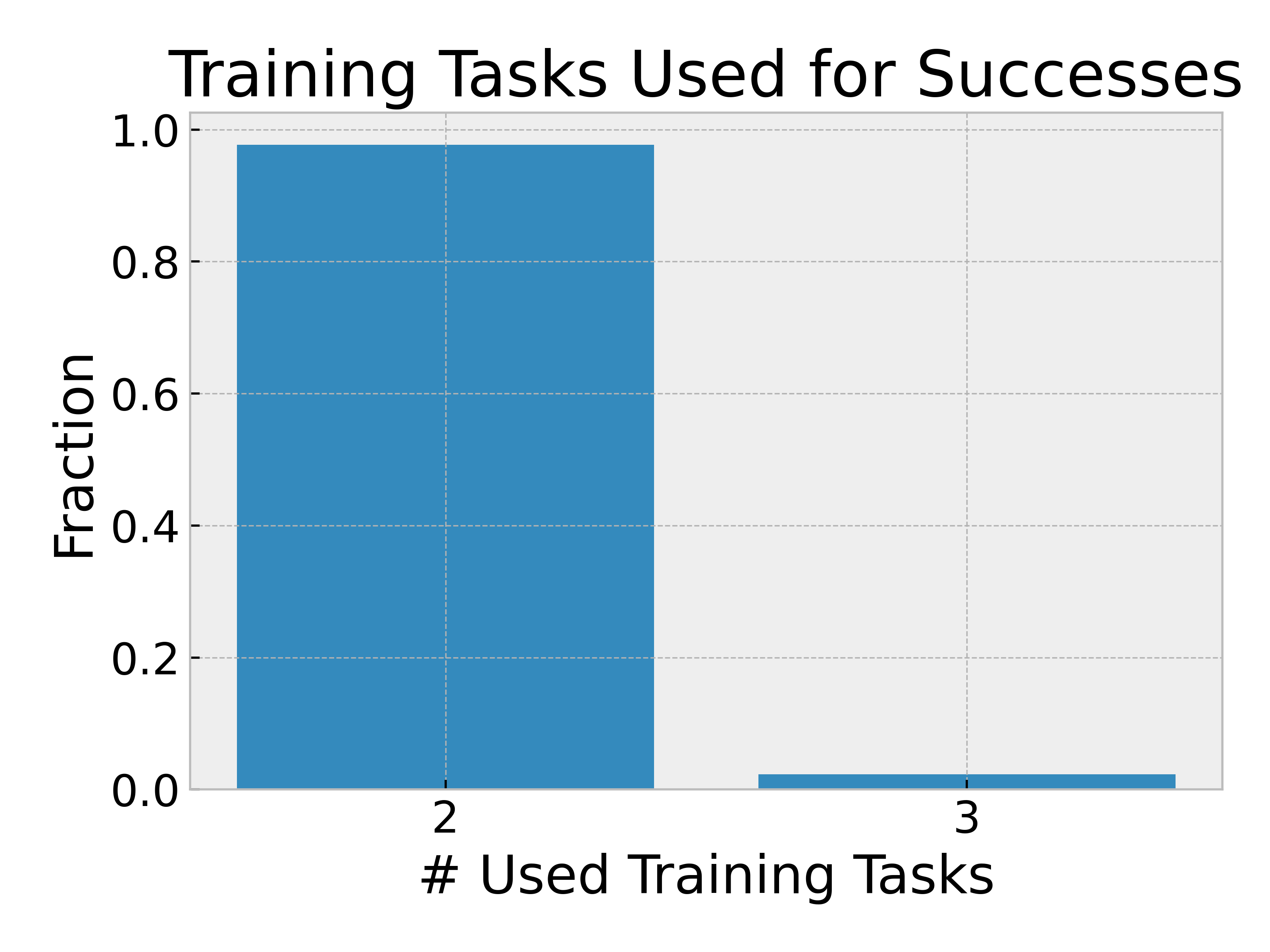}
    \caption{Data efficiency of GPT-4. The reported fractions are over all trials, over all domains.}
    \label{fig:num_training_tasks}
\end{figure}

Figure~\ref{fig:num_training_tasks} shows the number of training tasks \emph{used} in successful trials.
See main text for details.
In the vast majority of cases, only two tasks are necessary, indicating the strong data-efficiency of GPT-4 as a generalized planner.

\begin{table}[h]
\centering
\begin{tabular}{cccccc} 
 \toprule
 \small
 \textbf{Domain} & \small GPT-4 & \small  No CoT & \small  No Debug & \small No Names & \small GPT-3.5 \\
 \midrule
 \small Delivery & 1.00 & 1.00 & 1.00 & 1.00 & 0.00 \\
 \small Forest & 1.00 & 1.00 & 1.00 & 0.93 & 1.00 \\
 \small Gripper & 1.00 & 1.00 & 1.00 & 1.00 & 0.00 \\ 
 \small Miconic & 0.07 & 0.00 & 0.00 & 0.00 & 0.00 \\ 
 \small Ferry & 1.00 & 1.00 & 1.00 & 0.00 & 0.00 \\
 \small Spanner & 1.00 & 1.00 & 0.00 & 0.00 & 0.00 \\ 
 \small Heavy & 1.00 & 1.00 & 1.00 & 0.00 & 0.00 \\ 
 \bottomrule
\end{tabular}
\caption{Max fraction of evaluation tasks solved per seed.}
\label{table:max_results}
\end{table}

\section{Example Programs}

Examples of programs synthesized by GPT-4 with CoT summarization and automated debugging are shown in Listings~\ref{delivery_example1}, \ref{forest_example1}, \ref{gripper_example1}, \ref{miconic_example1}, \ref{ferry_example1}, \ref{spanner_example1}, and \ref{heavy_example1}.
Any comments in the code are added by GPT-4 itself.

\begin{listing*}[!ht]
\small
\begin{minted}{python}
def get_plan(objects, init, goal):
    locs = [obj[0] for obj in objects if obj[1] == "loc"]
    papers = [obj[0] for obj in objects if obj[1] == "paper"]
    
    home_base = next(loc for loc in locs if ("ishomebase", loc) in init)
    safe_locs = [loc for loc in locs if ("safe", loc) in init]
    wants_paper = [loc for loc in locs if ("wantspaper", loc) in init]
    
    plan = []
    
    for loc in wants_paper:
        if loc in safe_locs:
            # Pick up a newspaper at the home base
            paper = papers.pop()
            plan.append(f"(pick-up {paper} {home_base})")
            
            # Move to the safe location that wants a newspaper
            plan.append(f"(move {home_base} {loc})")
            
            # Deliver the newspaper to the location
            plan.append(f"(deliver {paper} {loc})")
            
            # Move back to the home base
            plan.append(f"(move {loc} {home_base})")
    
    return plan
\end{minted}
\caption{\textbf{Delivery} domain example synthesized program (success)}
\label{delivery_example1}
\end{listing*}

\begin{listing*}[!ht]
\small
\begin{minted}{python}
def get_plan(objects, init, goal):
    def find_location(init):
        for atom in init:
            if atom[0] == "at":
                return atom[1]
        return None

    def find_adjacent_trail(current_loc, init):
        for atom in init:
            if atom[0] == "ontrail" and atom[1] == current_loc:
                return atom[2]
        return None

    def is_hill(loc, init):
        for atom in init:
            if atom[0] == "ishill" and atom[1] == loc:
                return True
        return False

    current_loc = find_location(init)
    goal_loc = find_location(goal)
    plan = []

    while current_loc != goal_loc:
        next_loc = find_adjacent_trail(current_loc, init)

        if is_hill(next_loc, init):
            action = f"(climb {current_loc} {next_loc})"
        else:
            action = f"(walk {current_loc} {next_loc})"

        plan.append(action)
        current_loc = next_loc

    return plan
\end{minted}
\caption{\textbf{Forest} domain example synthesized program (success)}
\label{forest_example1}
\end{listing*}

\begin{listing*}[!ht]
\small
\begin{minted}{python}
def get_plan(objects, init, goal):
    def get_objects_of_type(type_name):
        return [obj for obj in objects if obj.startswith(type_name)]

    def get_location(state, ball):
        for atom in state:
            if atom[0] == 'at' and atom[1] == ball:
                return atom[2]
        return None

    balls = get_objects_of_type('ball')
    rooms = get_objects_of_type('room')
    grippers = get_objects_of_type('gripper')

    robby_location = None
    for atom in init:
        if atom[0] == 'at-robby':
            robby_location = atom[1]
            break

    plan = []
    for ball in balls:
        initial_location = get_location(init, ball)
        goal_location = get_location(goal, ball)

        if goal_location is None:
            continue

        if initial_location != goal_location:
            plan.append(f'(move {robby_location} {initial_location})')
            robby_location = initial_location

            plan.append(f'(pick {ball} {initial_location} {grippers[0]})')

            plan.append(f'(move {robby_location} {goal_location})')
            robby_location = goal_location

            plan.append(f'(drop {ball} {goal_location} {grippers[0]})')

    return plan
\end{minted}
\caption{\textbf{Gripper} domain example synthesized program (success)}
\label{gripper_example1}
\end{listing*}

\begin{listing*}[!ht]
\small
\begin{minted}{python}
def get_plan(objects, init, goal):
    floors_by_building = {}
    passengers_by_building = {}

    for obj, obj_type in objects:
        if obj_type == "floor":
            building = obj.split("_")[-1]
            if building not in floors_by_building:
                floors_by_building[building] = []
            floors_by_building[building].append(obj)
        elif obj_type == "passenger":
            building = obj.split("_")[-1]
            if building not in passengers_by_building:
                passengers_by_building[building] = []
            passengers_by_building[building].append(obj)

    for building in floors_by_building:
        floors_by_building[building] = sorted(floors_by_building[building])

    origins = {p: f for p, f in (atom[1:] for atom in init if atom[0] == "origin")}
    destins = {p: f for p, f in (atom[1:] for atom in init if atom[0] == "destin")}
    lift_at = {atom[1]: atom for atom in init if atom[0] == "lift-at"}

    plan = []
    boarded = set()

    for building, floors in floors_by_building.items():
        passengers = passengers_by_building[building]
        current_lift_at = lift_at[building]

        for floor in floors + list(reversed(floors)):
            if current_lift_at != floor:
                if current_lift_at < floor:
                    plan.append(f"(up {current_lift_at} {floor})")
                else:
                    plan.append(f"(down {current_lift_at} {floor})")
                current_lift_at = floor

            for passenger in passengers:
                if origins.get(passenger) == floor:
                    plan.append(f"(board {floor} {passenger})")
                    boarded.add(passenger)

                if destins.get(passenger) == floor and passenger in boarded:
                    plan.append(f"(depart {floor} {passenger})")
                    boarded.remove(passenger)

    return plan
\end{minted}
\caption{\textbf{Miconic} domain example synthesized program (failure)}
\label{miconic_example1}
\end{listing*}

\begin{listing*}[!ht]
\small
\begin{minted}{python}
def get_plan(objects, init, goal):
    plan = []

    # Helper function to find a car's location
    def get_location(car, state):
        for atom in state:
            if atom[0] == 'at' and atom[1] == car:
                return atom[2]
        return None

    cars = [obj for obj in objects if obj.startswith('c')]
    ferry_location = [atom[1] for atom in init if atom[0] == 'at-ferry'][0]

    for car in cars:
        initial_location = get_location(car, init)
        goal_location = get_location(car, goal)

        # Sail to the car's initial location
        if ferry_location != initial_location:
            plan.append(f'(sail {ferry_location} {initial_location})')
            ferry_location = initial_location

        # Board the car onto the ferry
        plan.append(f'(board {car} {initial_location})')

        # Sail to the car's goal location
        if ferry_location != goal_location:
            plan.append(f'(sail {ferry_location} {goal_location})')
            ferry_location = goal_location

        # Debark the car at the goal location
        plan.append(f'(debark {car} {goal_location})')

    return plan
\end{minted}
\caption{\textbf{Ferry} domain example synthesized program (success)}
\label{ferry_example1}
\end{listing*}

\begin{listing*}[!ht]
\small
\begin{minted}{python}
def get_plan(objects, init, goal):
    plan = []
    bob_location = get_bob_location(init)
    spanner_locations = get_spanner_locations(init)
    nuts_to_tighten = {n for pred, *args in goal if pred == 'tightened' for n in args}

    while nuts_to_tighten:
        closest_spanner = None
        min_distance = float('inf')

        for loc, spanner in spanner_locations.items():
            distance = get_shortest_path(init, bob_location, loc)
            if distance < min_distance:
                min_distance = distance
                closest_spanner = spanner
                closest_spanner_loc = loc

        # Move Bob to the closest spanner
        path_to_spanner = get_path(init, bob_location, closest_spanner_loc)
        for next_location in path_to_spanner:
            plan.append(f'(walk {bob_location} {next_location} bob)')
            bob_location = next_location

        # Pick up the spanner
        plan.append(f'(pickup_spanner {closest_spanner_loc} {closest_spanner} bob)')

        # Move Bob to the gate
        gate = 'gate'
        path_to_gate = get_path(init, bob_location, gate)
        for next_location in path_to_gate:
            plan.append(f'(walk {bob_location} {next_location} bob)')
            bob_location = next_location

        # Tighten a nut using the spanner
        nut = nuts_to_tighten.pop()
        plan.append(f'(tighten_nut {gate} {closest_spanner} bob {nut})')

        # Remove the used spanner from the available spanners
        if closest_spanner_loc in spanner_locations:
            del spanner_locations[closest_spanner_loc]

    return plan

def get_path(init, start, end):
    paths = [start]
    while paths[-1] != end:
        for pred, l1, l2 in init:
            if pred == 'link' and paths[-1] == l1 and l2 not in paths:
                paths.append(l2)
                break
    return paths[1:]
\end{minted}
\caption{\textbf{Spanner} domain example synthesized program (failure)}
\label{spanner_example1}
\end{listing*}

\begin{listing*}[!ht]
\small
\begin{minted}{python}
def extract_heavier_relations(init):
    heavier_relations = set()
    for atom in init:
        if atom[0] == 'heavier':
            heavier_relations.add(atom)
    return heavier_relations

def sort_objects_by_weight(objects, heavier_relations):
    sorted_objects = []
    while len(objects) > 0:
        for obj in objects.copy():
            if all(('heavier', other, obj) not in heavier_relations for other in objects):
                sorted_objects.append(obj)
                objects.remove(obj)
    return sorted_objects

def get_plan(objects, init, goal):
    heavier_relations = extract_heavier_relations(init)
    sorted_objects = sort_objects_by_weight(set(objects), heavier_relations)

    plan = []
    first_object = sorted_objects.pop(0)
    plan.append(f"(pack-first {first_object})")

    for obj in sorted_objects:
        plan.append(f"(stack {first_object} {obj})")
        first_object = obj

    return plan
\end{minted}
\caption{\textbf{Heavy} domain example synthesized program (success)}
\label{heavy_example1}
\end{listing*}

\end{document}